\documentclass[graybox]{svmult}

\usepackage{type1cm}        
\usepackage{makeidx}         
\usepackage{graphicx}        
\usepackage{multicol}        
\usepackage[bottom]{footmisc}

\usepackage{newtxtext}       %
\usepackage{newtxmath}       

\usepackage{algorithm}
\usepackage{algorithmic}
\usepackage[square,numbers]{natbib}
\usepackage{booktabs}
\usepackage{multirow}

\usepackage{fancyhdr}

\fancypagestyle{Whatever}{
  \fancyhf{} 
  
  \fancyhead[R]{\small\thepage}
  \fancyfoot[R]{\small DISTRIBUTION STATEMENT A. Approved for public release: distribution unlimited, AFRL-2021-3176.} 

}

\makeindex             

\begin{document}

\pagestyle{Whatever}
\title*{Homography augumented momentum constrastive learning for SAR image retrieval}
\author{Seonho Park, Maciej Rysz, Kathleen M. Dipple and Panos M. Pardalos}
\institute{Seonho Park \at Department of Industrial and Systems Engineering, University of Florida, Gainesville, FL. \email{seonhopark@ufl.edu}
\and Maciej Rysz \at Department of Information Systems \& Analytics, Miami University, Oxford, OH. \email{ryszmw@miamioh.edu}
\and Kathleen M. Dipple \at Integrated Sensor and Navigation Services Team, Air Force Research Laboratory (AFRL/RWWI), Eglin Air Force Base, FL. \email{kathleen.dipple.1@us.af.mil}
\and Panos M. Pardalos \at Department of Industrial and Systems Engineering, University of Florida, Gainesville, FL. \email{pardalos@ise.ufl.edu}}

%
%
\maketitle
\abstract{
Deep learning-based image retrieval has been emphasized in computer vision.
Representation embedding extracted by deep neural networks (DNNs) not only aims at containing semantic information of the image, but also can manage large-scale image retrieval tasks.
In this work, we propose a deep learning-based image retrieval approach using homography transformation augmented contrastive learning to perform large-scale synthetic aperture radar (SAR) image search tasks.
Moreover, we propose a training method for the DNNs induced by contrastive learning that does not require any labelling procedure. 
This may enable tractability of large-scale datasets with relative ease.
Finally, we verify the performance of the proposed method by conducting experiments on the polarimetric SAR image datasets.
}


\section{Introduction}\label{sec:intro}
Deep learning has been applied for synthetic aperture radar (SAR) image analysis tasks such as object detection \cite{chen2014sar}, despeckling \cite{tang2019sar,chierchia2017sar,cozzolino2020nonlocal}, optical data fusion \cite{schmitt2018sen1}, and terrain surface classification \cite{parikh2020classification}.
SAR images can capture the geographic characteristics without depending on the weather conditions.
One of the applications is the SAR image retrieval task \cite{noh2017large,teichmann2019detect,detone2018superpoint}, which aims to retrieve images from a large database that are similar to a query image.
This application is further utilized for complementing  navigation systems when global positioning system (GPS) is not available \cite{park2021synthetic}.
For the SAR image retrieval tasks, it is necessary to extract a compressed feature vector from a SAR image while maintaining the semantic information. The vector is then compared with the vectors of SAR images in the database.
This technique is sometimes called the global descriptor approach.
Convolutional neural networks (CNNs) are frequently used for extracting the global descriptor vectors from the images \cite{gordo2016deep,arandjelovic2016netvlad,radenovic2016cnn}.
The global descriptor is a simple vector with a prescribed dimension, thus during testing, distance between vectors can be easily and scalably measured.
However, this technique's overall performance may suffer from complications such as clutter, illumination, and occlusion that all hinder the CNNs from generating adequate global descriptor vectors.
To overcome these obstacles, the CNN-based local feature approaches are also suggested \cite{yi2016lift,noh2017large,teichmann2019detect,cao2020unifying,detone2018superpoint}.
These methods generate so called \textit{keypoints} and their local descriptors.
The keypoint represents the location of interest and the local descriptor is a vector characterizing an associated keypoint.
The CNN-based local feature approaches generally aim to replace the traditional local feature approaches, such as SIFT \cite{lowe1999object,lowe2004distinctive} or its variants \cite{dellinger2014sar}.
Even though this enhances the performance of retrieving images, comparing image pairs is not scalable, thus it cannot be efficiently applied over the large-scale database.
Thus, recent efforts usually focus on developing a CNN-based apporoach combining both global descriptor and local feature approaches; first retrieving images roughly using global descriptor and reranking by utilizing the local feature approach \cite{noh2017large,park2021synthetic}.

In this work, we focus only on the CNN based global descriptor method, which could be a main principle component of the deep learning-based SAR image retrieval system.
We propose contrastive learning \cite{hadsell2006dimensionality} for generating the global descriptor.
Contrastive learning uses multiple neural networks and compares the global descriptors relative to a loss function.
To prevent the networks from generating trivial descriptors, homography transformation is applied to augmenting the SAR images.
We also demonstrate that the homography transformation can generalize SAR image deformation, thus the homography transformed SAR images are used as an input data for training the networks.
Contrastive learning enables the network to learn the features in a self-supervised way, which potentially leads to dealing with a large-scale dataset without involving any arduous labelling process usually done by human labor.
To verify the performance of the proposed method, we show experiment results on multiple SAR datasets containing various geographic characteristics.

This paper is organized as follows. 
Related works of our approach are given in Section \ref{sec:related_works}. 
In Section \ref{sec:methodology}, a method for generating global descriptor is proposed, which includes contrastive learning and homography transformation.
The performance results of the proposed approach on the public SAR dataset are reported in Section \ref{sec:experiments}.
Finally, Section \ref{sec:conclusions} presents the conclusions.
\section{Related Works}\label{sec:related_works}
\paragraph{\textit{Contrastive learning}}
Contrastive learning \cite{hadsell2006dimensionality} is a self-supervised learning method which has been actively researched recently.
It primarily utilizes a comparison between the feature representation pairs of the instances in a representative space to form a loss function.
Contrastive learning stems from the idea that the feature vectors of the image and its augmented image are to be located closely while the two feature vectors of two distinguishable images should be located far off.
Generally, contrastive learning enables the machine learning system not to rely on labels, but instead, it leads to learning with the \textit{pseudo labels} generated automatically while comparing the feature vectors during training.
In regards to comparing between vectors, 
it collects vectors from previous iterations and saves them in a memory storage.
Contrastive learning can be also utilized for pretraining in a self-supervised manner in order to enhance the performance before applying it for downstream tasks \cite{chen2021empirical}.
The primitive attempt, called a \textit{memory bank} scheme, stores the whole vectors obtained at the previous iterations and uses a subset of them at the current iteration \cite{wu2018unsupervised,park2021synthetic}. 
Since the encoder is gradually updated via backpropagation and a stochastic gradient descent algorithm, the output vectors stored in the memory bank are on occasion incompatible with those produced at the current iteration, which, in turn, leads to slow learning.
Further, it involves memory issues due to the notoriously high number of the training data instances.
To circumvent this, Momentum Contrast (MoCo) \cite{he2020momentum} uses the Siamese encoder, that is, there are two distinguishable encoders with the same architectures from which the one, primary encoder, is updated with the gradient; whereas the other, momentum encoder, is updated using a momentum rule.
Also, it stores the restricted numbers of the \textit{key} vectors outputted by the momentum encoder into a queue matrix.
These are then used when comparing with a \textit{query} vector outputted by the primary encoder.

There also exist techniques to improve performance of the MoCo mechanism including adding a 2-layer MLP head, blur augmentation, and cosine learning rate schedule \cite{chen2020improved}.
Recent improvements of MoCo, namely MoCo v3 \cite{chen2021empirical}, suggested an online method generating keys in the same batch to reduce the memory uses while maintaining fast learning.
It also uses a form of the infoNCE \cite{oord2018representation} as a loss function allowing for representative differentiations between images of a given sufficient batch size (e.g., 4096).
Additionally, to further improve the performance of MoCo, they adopted a vision transformation (ViT) \cite{dosovitskiy2020image} as the backbone architecture instead of CNN.
There are various approaches in contrastive learning 
that differ depending on the usages of the queue matrix and the forms of the loss function.
They include SimCLR \cite{chen2020simple}, SEER \cite{goyal2021self}, SwAV \cite{caron2020unsupervised}, SimSiam \cite{chen2020exploring}, and BYOL \cite{grill2020bootstrap}.

\paragraph{\textit{Deep learning-based image retrieval}}
Deep neural networks (DNNs) have been widely used for representative learning.
Traditional components of image retrieval systems, such as SIFT \cite{lowe1999object,lowe2004distinctive}, RANSAC \cite{fischler1981random}, have been partially replaced by DNN based approaches.

Noh et al. \cite{noh2017large} introduced a method for generating local features using the CNN based model, which can substantially replace the traditional local descriptors such as SIFT.
The central point of the receptive field corresponding to the local descriptor represents the keypoint, and the local descriptor is trained with the attention score layer.
They also proposed a landmark dataset on which the proposed model is finetuned with the location based image classification labels, which can be regarded as a weakly supervised approach.
The performance of this method is further enhanced by applying the regional aggregated selective match kernels \cite{teichmann2019detect} and generating both the global descriptor and local features simultaneously \cite{cao2020unifying}.
SuperPoint \cite{detone2018superpoint} outputs both the keypoints and local descriptors simultaneously via a unified CNN-based backbone architecture.
Its training process involves pretraining on a synthetic labeled dataset and self-supervised learning with the help of homography transformation.

Park et al. \cite{park2021synthetic} proposed deep cosine similarity neural network to generate a $l2$ normalized feature vector of a SAR image with the primary purpose of comparing SAR image pairs scalably.
The proposed idea of normalizing the vector during training is also equipped in the training process of the encoders used in the present work so as to maintain its norm consistently.

Once similar images from a database are retrieved using the global descriptors, the local features are used for reranking the retrieved images.
This process is usually time consuming since matching pairs through RANSAC \cite{fischler1981random} or Direct Linear Transformation (DLT) \cite{andrew2001multiple} is not scalable.
Thus, replacing these matching techniques with scalable DNN-based methods is in an active research area where many approaches such as SuperGlue \cite{sarlin2020superglue} and LoFTR \cite{sun2021loftr} have been proposed.
They also utilize a state-of-the-art architecture, Transformer \cite{vaswani2017attention}, as their backbone networks, which  can significantly enhance performances of the DNN based methods for image retrieval tasks.

\section{Methodology}\label{sec:methodology}
We generate a vector, the (global) feature vector $\vec{d}$, to compress and represent a SAR image for the purpose of SAR image retrieval. 
Namely, the vector contains semantic information of the SAR image. In this section, we focus on training a deep neural network based encoder to produce a feature vector $\vec{d}$ that is used in the SAR image retrieval task. 


\subsection{Contrastive Learning}
Contrastive learning has been emphasized recently in the literature as a means for generating feature embedding in a self-supervised way.
Generally, it uses the Siamese encoder (twin encoders) that consists of two distinguished CNN based encoders with the same architectures and the same sets of parameters to be learned separately.
When two inputs of two different encoders are similar, contrastive learning aims at producing output vectors that are likewise similar; whereas when two inputs are not similar to each other, it should produce outputs that are distinguishable.
Contrastive learning allows the encoder to be trained in a self-supervised way; the input data does not need to use labels during training, thus one can aggregate more data for training the model without involving any arduous labor commonly required for labelling instances.

\begin{figure*}[ht!]
\centering
\includegraphics[width=0.7\textwidth]{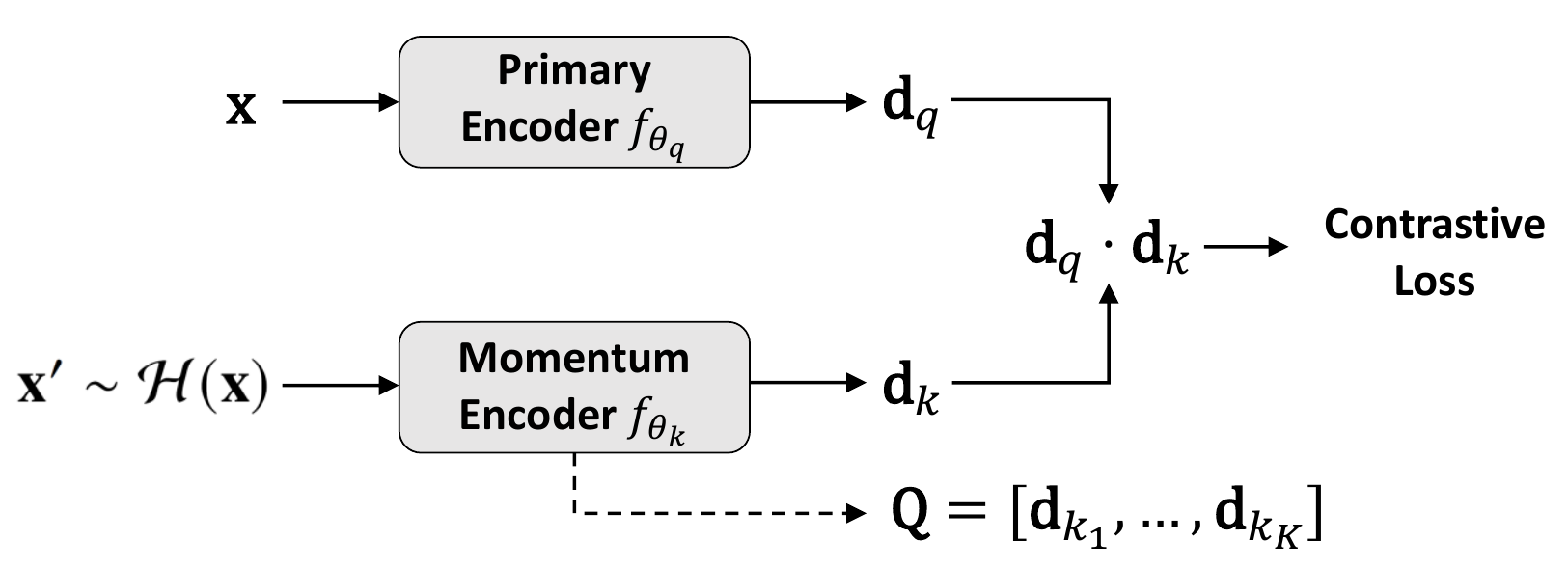} 
\caption{Diagram of the contrastive learning}\label{fig:contra}
\vskip 0.2in
\end{figure*}

In this work, we use MoCo-like learning \cite{he2020momentum} where the first encoder is updated via a stochastic gradient descent algorithm, whereas the second encoder is updated by a momemtum rule.
Fig. \ref{fig:contra} depicts the contrastive learning framework that we use for SAR image retrieval.
As shown in the figure, two CNN based encoders, the ``primary encoder'' $f_{\vec{\theta}_q}$ and the ``momentum encoder'' $f_{\vec{\theta}_k}$, are deployed, where $\vec{\theta}_q$ and $\vec{\theta}_k$ are the parameter sets corresponding to the former and latter, respectively.

The primary encoder $f_{\vec{\theta}_q}$ uses a SAR image as an input $\vec{x}$ to produce an output $\vec{d}_q$. 
Similarly, the momentum encoder $f_{\vec{\theta}_k}$, which has the same architecture as the primary encoder, uses a transformed SAR image $\vec{x}'\sim\mathcal{H}(\vec{x})$ that is drawn from homography transformation based distribution $\mathcal{H}(\vec{x})$ to output a feature vector $\vec{d}_k$.
The primary encoder is updated using a stochastic gradient descent method where the gradients associated with the parameters are computed through backpropagation.
The loss function used for training the primary encoder is described in subsection \ref{subsec:training}.
The parameters $\vec{\theta}_k$ in the momentum encoder are updated with the momentum rule $\vec{\theta}_k \leftarrow m\vec{\theta}_k + (1-m)\vec{\theta}_q$, where $m$ is a momentum parameter used to slowly and stably update parameters $\vec{\theta}_k$.
Also, in the momentum encoder,  
the queue matrix $\vec{Q}$ consists of columns of feature vectors calculated during the previous iterations: 

\begin{equation}
    \vec{Q} = \left[\vec{d}_{k_1},\dots,\vec{d}_{k_K}\right],
\end{equation}
where $K$ is a prescribed parameter representing the number of feature vectors. 

Matrix $\vec{Q}$ is continuously enqueued using the most recently generated vectors at the current iteration while the oldest vectors are dequeued to maintain the size of the matrix. 
It is also emphasized that the momentum encoder uses the transformed SAR image as an input and in this work we consider homography transformation for transforming the SAR images.

\subsection{Homography transformation}
The transformed input $\vec{x}'$ used by the momentum encoder should maintain the same semantic information as the original image, and simultaneously
generalize the images in order to prevent trivial training cases. 
To this end, in the present endeavor we utilize homography transformation. 
It is known that homography transformation can explain the translational and rotational transformations between an image pair in a same planar space.
Thus, it is suitable for tasks such as the one of present interests where SAR images are taken from aircrafts or satellites that are positioned at various altitudes and angles to a specific surface area 
on a given planar space. 
Given a pixel location $[u,v]$ in an original image, homography transformation is a linear transformation represented by a non-singular $3\times3$ matrix as

\begin{equation}\label{eq:homo}
    \begin{bmatrix}
    u'\\
    v'\\
    1
    \end{bmatrix}
    =
    \begin{bmatrix}
    H_{11}&H_{12}&H_{13}\\
    H_{21}&H_{22}&H_{23}\\
    H_{31}&H_{32}&H_{33}\\
    \end{bmatrix}
    \begin{bmatrix}
    u\\
    v\\
    1
    \end{bmatrix},\,
    \vec{u}' = \vec{H}\vec{u},
\end{equation}
where $H_{ij},\,\forall i,j\in\{1,2,3\}$ represents the element of the homography transformation matrix $\vec{H}$, and $[u',v']$ is the transformed location.
Note that the homography transformation matrix is invariant with scaling, thus containing 8 degrees of freedom.
To augment the image data via homography transformation, a 4-point based homography transformation 
is frequently
utilized \cite{detone2016deep,nguyen2018unsupervised}.
We employ a similar approach furnished in \cite{detone2016deep} to generate the homographic transformed SAR image $\vec{x}'$, which will serve as an input to the momentum encoder.

\begin{figure*}[ht!]
\centering
\includegraphics[width=0.7\textwidth]{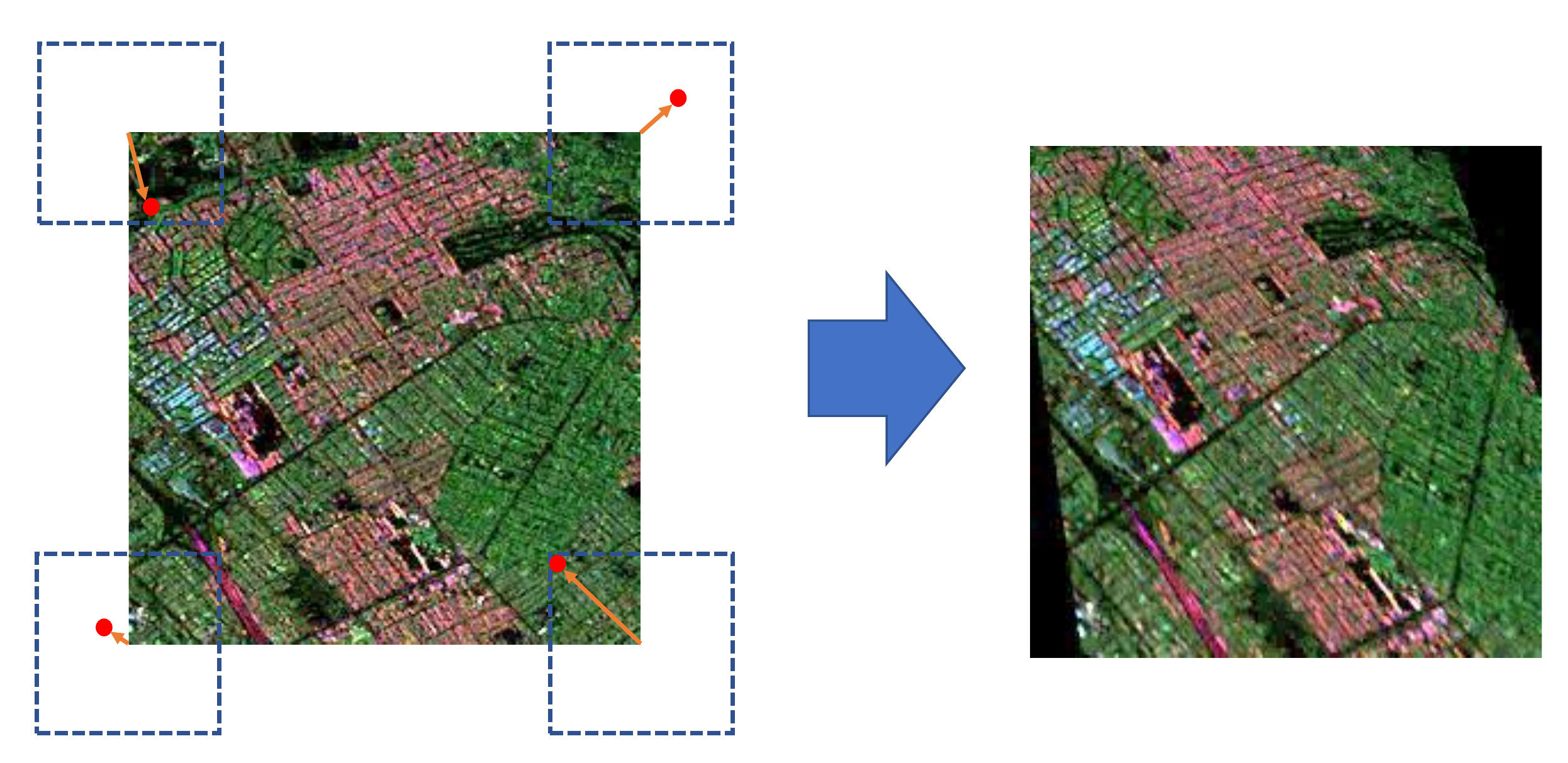} 
\caption{An example of 4-points based homography tranformation for SAR images}\label{fig:homo}
\vskip 0.2in
\end{figure*}

As shown in Fig. \ref{fig:homo},  observe that the four corners of the pixel points $(u_k,v_k)$ for $k\in\{1,2,3,4\}$ (represented as red dots) are drawn at random within the small blue squares. 
Then, using the four corner points we construct a 4-point parameterization $\vec{H}_{4p}$ that consists of the $x$, $y$ displacements, $(\Delta u_k=u_k'-u_k, \Delta v_k=v_k'-v_k)$ for $k\in\{1,2,3,4\}$.
Note that the 4-point parameterization $\vec{H}_{4p}$ is a $4\times2$ matrix containing 8 dofs as the homography transformation matrix $\vec{H}$, and that  $\vec{H}_{4p}$ can easily be converted to $\vec{H}$ by the DLT algorithm \cite{andrew2001multiple} or the function \texttt{getPerspectiveTransform} in OpenCV.
As shown in the right side of Fig. \ref{fig:homo},
after applying the homography transformation to an image, 
the transformed image is cropped so that it is of the same size as the original image.
In what follows, with a slight abuse in notation we denote the image distribution of the homography transformation and cropping applied to the image $\vec{x}$ as $\mathcal{H}(\vec{x})$.
The training procedure is presented next.

\subsection{Training}\label{subsec:training}

The transformed image $\vec{x}'\sim \mathcal{H}(\vec{x})$ based on $\vec{H}_{4p}$ is first sampled
and serves as input of the momentum encoder as 
\begin{align}
    \vec{d}_q = f_{\vec{\theta}_q}(\vec{x}),\\
    \vec{d}_k = f_{\vec{\theta}_k}(\vec{x'}),
\end{align}
while the original SAR image is passed as input to the primary encoder. After $l2$-normalizing feature vectors $\vec{d}_q$ and $\vec{d}_k$, a logit vector is constructed as follows:

\begin{equation}\label{eq:logit}
    \vec{l}_q = \left[\frac{\vec{d}_q^T\vec{d}_k}{\tau},\frac{\vec{d}_q^T\vec{Q}}{\tau}\right]^T    ,
\end{equation}
where parameter $\tau>0$ is called a \textit{temperature} parameter \cite{hinton2015distilling,pereyra2017regularizing}. 
For each SAR image $\vec{x}$ in the training dataset, the loss function takes the form of the infoNCE \cite{oord2018representation}, i.e.,

\begin{equation}
    L(\vec{x}) = -\log \frac{e^{l_{q_1}}}{\sum_{i=1}^{K+1}e^{l_{q_i}}},
\end{equation}
where $l_{q_i}$ represents the $i$th element of the logit vector $\vec{l}_q$.
Finally, for the training dataset $\{\vec{x}^{(1)},\dots,\vec{x}^{(N)}\}$, the overall loss function for training the primary encoder is given by

\begin{equation}\label{eq:loss}
    \min_{\vec{\theta}_q} L = \frac{1}{N}\sum_{i=1}^{N}L(\vec{x}^{(i)}).
\end{equation}

The procedure for training the primary encoder and its corresponding momentum encoder is presented in Algorithm \ref{alg:algorithm}.
As shown in line 9 in Algorithm 1, backpropagation is conducted to calculate the gradients of the loss function w.r.t. the parameters $\vec{\theta}_{q}$.
Stochastic gradient descent or its variants \cite{kingma2014adam,park2020combining} can be applied as $\text{Opt}(\vec{\theta}_q, \eta, \Delta\vec{\theta}_q)$ to update the parameters $\vec{\theta}_{q}$, where $\Delta\vec{\theta}_q$ represents the gradients or their variants used for updating the parameters.

\begin{algorithm}[tb]
  \caption{Contrastive learning of feature vector for SAR image retrieval}
  \label{alg:algorithm}
\begin{algorithmic}[1]
    \STATE Input: learning rate $\eta$, minibatch size $N$, queue size $K$, momentum parameter $m$, temperature parameter $\tau$
    \STATE Initialize queue matrix $\vec{Q}$, parameters of the encoders as $\vec{\theta}_q=\vec{\theta}_k$
    \REPEAT
    \STATE Draw $\vec{x}^{(1)},\dots,\vec{x}^{(N)}$ samples at random from the training dataset
    \STATE Generate homography transformed images $\vec{x}'^{(1)},\dots,\vec{x}'^{(N)}$
    \STATE Calculate $\vec{d}_q^{(i)} = f_{\vec{\theta}_q}(\vec{x}^{(i)})$ for all $i$
    \STATE Calculate $\vec{d}_k^{(i)} = f_{\vec{\theta}_k}(\vec{x}'^{(i)})$ for all $i$
    \STATE Calculate loss function $L$ (Eq. \ref{eq:loss})
    \STATE Update $\vec{\theta}_q$ with $\text{Opt}(\vec{\theta}_q, \eta, \Delta\vec{\theta}_q)$
    \STATE Update $\vec{\theta}_k$ as $\vec{\theta}_k\leftarrow m\vec{\theta}_k+(1-m)\vec{\theta}_q$
    \STATE dequeue old $N$ columns from $\vec{Q}$
    \STATE enqueue new $N$ columns to $\vec{Q}$
    \UNTIL{$\vec{\theta}_q$  has converged}
    \STATE Output: learned parameters $\vec{\theta}_q$
\end{algorithmic}
\end{algorithm}

\section{Experiments}\label{sec:experiments}
\subsection{SAR image data}
As an experimental testbed for SAR image retrieval, we utilize  the Uninhabited Aerial Vehicle Synthetic Aperture Radar (UAVSAR) images \cite{UAVSAR} from NASA.
From the UAVSAR database, we use L-band and the ground projected complex cross (GRD) products of the polarimetric SAR (PolSAR).
A central reason for using GRD products is that it provides information on pixel-wise geographic coordinates mapping that, in turn, leads to precise evaluation of performance of image retrieval.
Specifically, HHHH, HVHV, VVVV products are processed to grayscale images through ASF MapReady 3.2 software, and  grayscale images of HHHH, HVHV, VVVV products correspond to red, green, blue channels of the resulting PolSAR images, respectively.

\begin{table}[t]
\begin{center}
\caption{PolSAR image maps from UAVSAR database}\label{table:sardata}
\begin{tabular}{@{}ccccc@{}}
\toprule
Name                 & Acquisition Date  & Size (W$\times$H)         & Region                                         & Characteristic                       \\\midrule
Haywrd1 & 10/09/2018               & 19779$\times$26236 & \multirow{2}{*}{Hayward fault, CA}             & \multirow{2}{*}{Building, Mountain}  \\
Haywrd2 & 05/30/2019               & 19758$\times$26206 &                                                &                                      \\\midrule
ykdelB1 & 08/28/2018               & 23007$\times$3906  & \multirow{2}{*}{Yukon-Kuskokwim delta, AK} & \multirow{2}{*}{River delta, Tundra} \\
ykdelB2 & 09/17/2019               & 23107$\times$3904  &                                                &                                      \\\midrule
atchaf & 04/01/2021               & 6436$\times$7973   & Atchafalaya river delta, LA                    & River delta, Vegetation              \\\midrule
harvrd & 08/17/2009               & 5981$\times$9319   & Harvard forest, MA                             & Forest                               \\\midrule
SanAnd & 02/21/2019               & 43050$\times$6604  & LA basin, CA                 & Building, Mountain                   \\\midrule
SRIVER & 06/17/2017               & 16631$\times$21099 & Subarctic tundra area, AK                  & Tundra \\
\bottomrule
\end{tabular}
\end{center}
\end{table}

\begin{figure*}[ht!]
\centering
\includegraphics[width=0.95\textwidth]{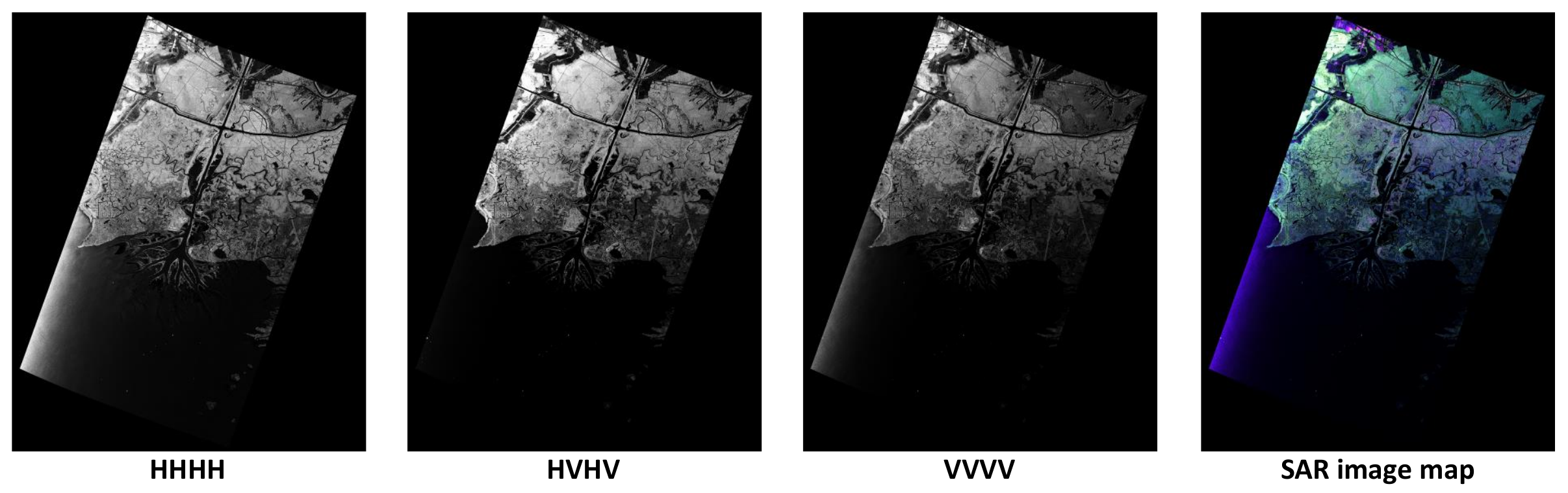} 
\caption{SAR image map of the 'atchaf' SAR data. RGB channels correspond to HHHH, HVHV, VVVV GRD products.}\label{fig:sarexample}
\vskip 0.2in
\end{figure*}

Table \ref{table:sardata} lists details of the SAR image maps used in our experiments that contain various topological characteristics including building, maintain, river, and vegetation.
As shown in the table, the SAR image maps are too big to serve as input for the encoders.
Thus, we extract patches of size $600\times600$ with a  stride of $100$ pixels from the images.
Since GRD formatting is arranged to locate the north to the upper side of an image map with respect to the geographic coordinates, it may contains a lot of ``blank'' area (shown as black region as in Fig. \ref{fig:sarexample}).
Further, as the blank area does not contain any meaningful information, we eliminate this by using the local descriptor method, SAR-SIFT \cite{dellinger2014sar}.
For each extracted patch, we generated the keypoints by SAR-SIFT and patches with at least 200 keypoints were added to the datasets in our experiments.
It was deemed that a patch with less than 200 keypoints contains a significant amount of black area.
The resulting patches were then resized to $224\times224$ pixels.
Homography transformation was applied ``on the fly'' during training. 
On the resized SAR patches, i.e., $224\times224$ pixels, four points were drawn at random, from the four corner squares of $32\times32$ pixels centered at the corner points (shown as the dashed blue lines in Fig. \ref{fig:homo}) to estimate $\vec{H}_{4p}$.

\begin{figure*}[ht!]
\centering
\includegraphics[width=0.6\textwidth]{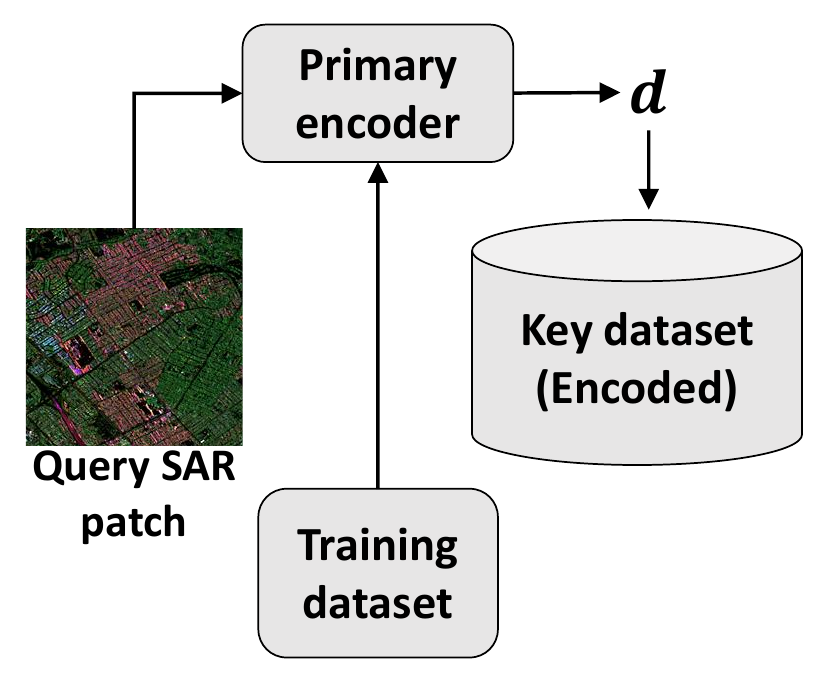} 
\caption{Diagram of SAR patch datasets usage}\label{fig:setdiagram}
\vskip 0.2in
\end{figure*}

For each experiment, three types of datasets consisting of the SAR patches were prepared: the query dataset, training dataset, and key dataset as shown in Fig. \ref{fig:setdiagram}.
Each of the patches in the datasets has at least 1000 SAR-SIFT based keypoints.
The primary encoder and momentum encoder are trained on the training dataset whereas the SAR patches in the query dataset are used for testing.
During testing, we extract the patches with similar feature vectors to the query patch from the key dataset which is also regarded as a database.
We restrict the number of patches in the query dataset to 100 which are selected at random.

\begin{table}[ht!]
\begin{center}
\caption{Details on experiment dataset. The number of SAR patches are shown in parenthesis.}\label{table:exp}
\begin{tabular}{@{}ccccc@{}}
\toprule
Exp. Num. & Exp. Name         & Query set     & Key set                  & Training set                                                                      \\ \midrule
Exp.1& Haywrd-Easy & Haywrd2 (100) & Haywrd1, ykdelB1 (20388) & Haywrd1 (12979)                                                                   \\ \midrule
Exp.2& Haywrd-Hard & Haywrd2 (100) & Haywrd1, ykdelB1 (20388) & \begin{tabular}[c]{@{}c@{}}atchaf, harvrd, \\ SanAnd, SRIVER (21555)\end{tabular} \\ \midrule
Exp.3& ykdelB-Easy & ykdelB2 (100) & Haywrd1, ykdelB1 (20388) & ykdelB1 (7409)                                                                    \\ \midrule
Exp.4& ykdelB-Hard & ykdelB2 (100) & Haywrd1, ykdelB1 (20388) & \begin{tabular}[c]{@{}c@{}}atchaf, harvrd, \\ SanAnd, SRIVER (21555)\end{tabular} \\ \bottomrule
\end{tabular}
\end{center}
\end{table}

To verify the performance of the proposed approach, we set up four experiments that simulate multiple operational circumstances, as shown in Table \ref{table:exp}.
The query sets are Haywrd2 or ykdelB2. 
Those are relatively recently acquired than the key datasets, Haywrd1 and ykdelB1 (Please see Table \ref{table:sardata}).
For experiment 1 and 3 (Exp.1 and Exp.3), which is harder experiments, we have used Haywrd1 and ykdelB1 as the training datasets, respectively. 
For these experiments, the training datasets were the same as the key datasets, hence the encoders have already observed similar SAR image patches to the query images during training.
It is therefore anticipated that the encoders would be able to generate more accurate feature vectors for these two experiments compared to the experiment cases 2 and 4 (Exp.2 and Exp.4).

For these easier experiments (Exp.2 and Exp.4), the encoders had not observed the same patches during training; instead, it was trained on the training dataset consisting of patches from atchaf, harvrd, SanAnd and SRIVER SAR maps.
Also for all experiments, the Haywrd1 and ykdelB1 are used as the key dataset.

\subsection{Architecture and Settings}

All experiments used ResNet50 or ResNet101 \cite{he2016deep} as a backbone CNN architecture.
The CNNs were pretrained on the ImageNet dataset \cite{deng2009imagenet}.
From the output map of the \textit{conv4} layer of ResNet, we adopted generalized mean (GeM) pooling \cite{radenovic2018fine} that weighs the importance of the outputs to generate the feature vector.
Denote that the output map of the \textit{conv4} layer as $\vec{o}_{h,w}$, where $h$ and $w$ are indices of height and width locations of the image, respectively.
We further integrated GeM pooling followed by a fully connected layer, parameterized by weights $F$ and bias $b_F$, similar to the approach in \cite{gordo2017end, cao2020unifying}.
The feature vector $\vec{d}$ that summarizes the discriminative outputs of the image is produced as an output of the GeM pooling layer as

\begin{equation}
    \vec{d} = F\times \left(\frac{1}{H_{\textit{conv4}}W_{\textit{conv4}}} \sum_{h,w}\vec{o}_{h,w}^p\right)^{1/p}+b_F,
\end{equation}
where $H_{\textit{conv4}}$, $W_{\textit{conv4}}$ represent the height and width of the output map, and $p$ is the parameter of the GeM pooling. In our experiments parameter$p$ is set to 3.
The feature vector $\vec{d}$ is then $l2$ normalized.

For training we employed the Adam optimizer \cite{kingma2014adam} with a learning rate of $5\mathrm{e}{-3}$ that gets decreased by $0.1$ after $80$ epochs.
The maximum number of epochs was set to 100 and the batch size $N$ for training is 32.
The temperature parameter $\tau$ was set to $0.5$ without performing any hyperparameter tunings.
For contrastive learning parameters, the queue size $K = 1024$ and the momentum parameter $m=0.999$.
To prevent feature vector $\vec{d}$ from having an excessively high $l2$ norm, prior to  
normalization we added the $l2$ regularization term derived in \cite{park2021synthetic} into the loss function.
Specifically, this term is simply denoted as $(||\vec{d}||_2-1)^2$.
The coefficient of the $l2$ regularization was set to $0.1$.

Note that when testing the image retrieval performance, since focus is put on measuring the performance of the feature vector, we do not consider any approximate techniques for measuring distance such as query expansion \cite{chum2007total,radenovic2018fine} or asymmetric quantizer distance \cite{cao2016deep}. 

\subsection{Experiments Results}

\begin{table}[ht!]
\begin{center}
\caption{Performance results from experiments}\label{table:results}
\begin{tabular}{l|rrrr}
\toprule
Method         & mAP & mP@1 & mP@10 & mP@50 \\\midrule
\multicolumn{5}{c}{\textbf{Exp.1 Haywrd-Easy}}       \\
ResNet50-512   & 0.5254 & 1.0000 &0.9870&0.8044\\
ResNet50-1024  &0.4908&1.0000&0.9810&0.7702\\
ResNet101-512  &0.5366&0.9700&0.9650&0.8018\\
ResNet101-1024\;\; &0.5314&0.9800&0.9630&0.7952\\\midrule
\multicolumn{5}{c}{\textbf{Exp.2 Haywrd-Hard}}       \\
ResNet50-512   &0.4092&1.0000&0.9810&0.6874\\
ResNet50-1024  &0.4287&1.0000&0.9730&0.7188\\
ResNet101-512  &0.4223&1.0000&0.9690&0.7050\\
ResNet101-1024 &0.4159&1.0000&0.9650&0.6976\\\midrule
\multicolumn{5}{c}{\textbf{Exp.3 ykdelB-Easy}}       \\
ResNet50-512   &0.4708&0.9700&0.9230&0.6128\\
ResNet50-1024  &0.4439&0.9100&0.8640&0.5760\\
ResNet101-512  &0.4296&0.9400&0.8690&0.5644\\
ResNet101-1024 &0.4290&0.9500&0.8630&0.5574\\\midrule
\multicolumn{5}{c}{\textbf{Exp.4 ykdelB-Hard}}       \\
ResNet50-512   &0.4181&0.9400&0.8440&0.5750\\
ResNet50-1024  &0.3916&0.9600&0.8340&0.5442\\
ResNet101-512  &0.3681&0.9700&0.8680&0.5138\\
ResNet101-1024 &0.3682&1.0000&0.8970&0.5280\\\bottomrule
\end{tabular}
\end{center}
\end{table}

To measure performance, we define precision and recall as follows.
Recall is the total number of the retrieved SAR patches from the database (key dataset), and
precision is defined as the ratio between the number of accurately retrieved SAR patches and the total number of the retrieved SAR patches.
When there are the duplicate regions calculated in terms of the geographic coordinates between the query image and the retrieved image, we consider the retrieved image as the accurately retrieved image.
As a primary performance measure, we have used mean average precision (mAP).
The average precision is the mean value of the precision values obtained at various recall values for each query image.
In the present setting, the mAP is the mean of the average precision values over 100 query images.
Also, the mean precision at $n$ (mP@n) is used as a performance measure where $n$ represents the recall value.
We report three mean precisions; mP@1, mP@10, mP@50.
For instance, it is implied that when we retrieve ten SAR patches from the database, mP@10 represents the fraction of the number of the accurate images we can expect to retrieve correctly.

The experiment results are shown in Table \ref{table:results}.
The methods vary depending on the dimension of the feature output $\vec{d}$, which is 512 or 1024, and the two considered architectures ResNet50 or ResNet101.

The ``Haywrd'' cases (Exp.1 and Exp.2) are from the Hayward fault in California, USA, and predominantly consists of human-made structures.
The ``ykdelB'' cases (Exp.3 and Exp.4) on the other hand, mostly consist of natural formations. 
By comparing the mAP values, we observed that the proposed method more effectively retrieved similar images for the Haywrd experiment cases compared to the ykdelB cases.
For both datasets, it can be seen that mAP values of the ``Hard'' cases are lower than those of the ``Easy'' cases.

Although the performances are relatively lower in the hard cases, 
it can be observed that the proposed method works well regardless of the dimension of the feature vector and the backbone architectures, as even mP@1 results in high values (over 0.9) in all the experiments. This result strongly indicates that our method can perform remarkably well without the use of any sophisticated yet non-scalable methods such as local descriptors \cite{noh2017large} and matching through RANSAC \cite{fischler1981random}.

\begin{figure*}[ht!]
\centering
\includegraphics[width=0.95\textwidth]{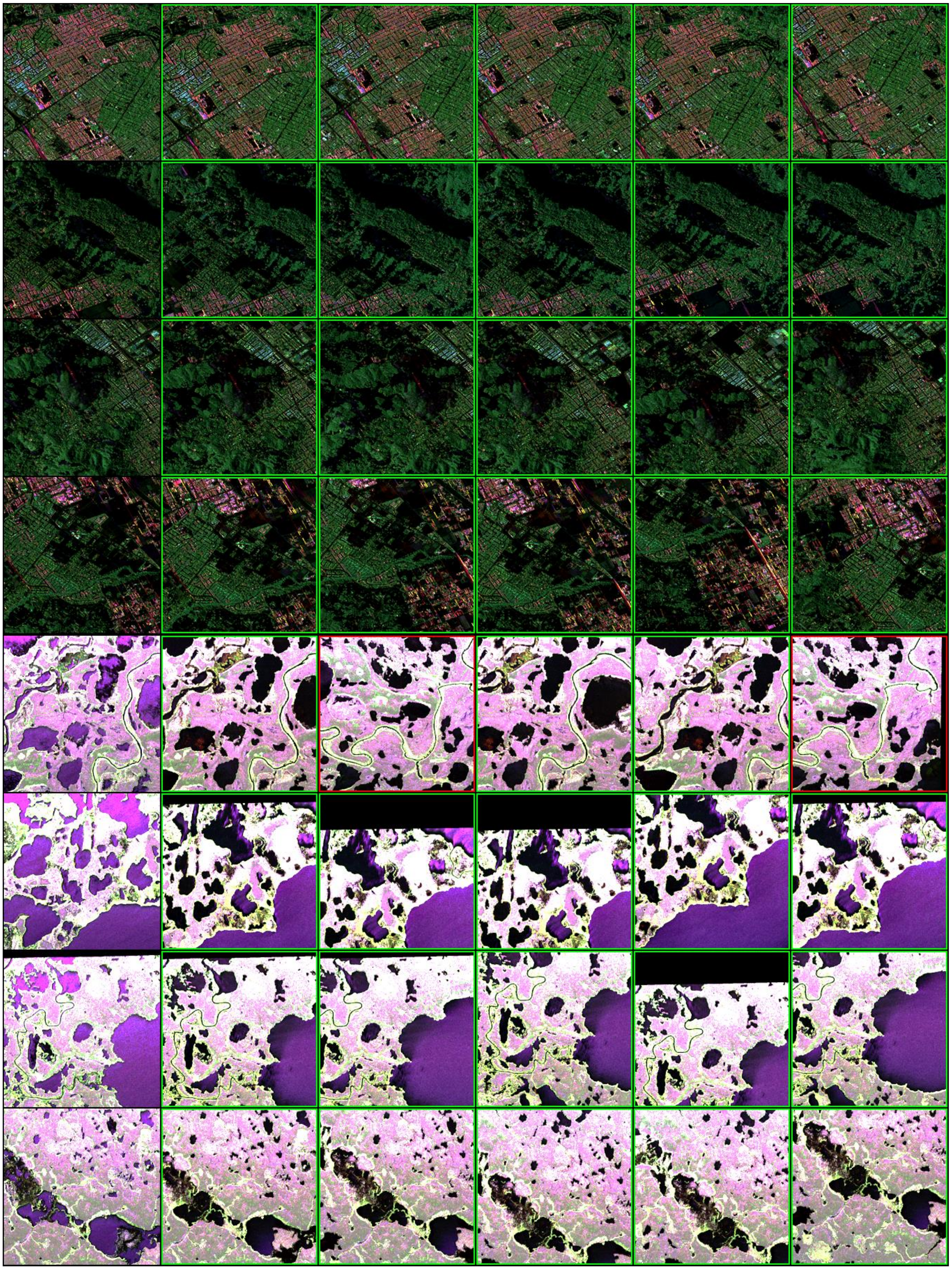} 
\caption{The retrieved SAR image examples. The first column is the query SAR images and others are the retrieved ones (the second column is the top retrieved and rightmost column is fifth retrieved). The green box represents an accurately retrieved SAR image, whereas the red box presents a incorrectly retrieved one. \textbf{Top}: Exp.2 Haywrd-Hard \textbf{Bottom}: Exp.4 ykdelB-Hard}\label{fig:exampleretrieve}
\vskip 0.2in
\end{figure*}

By way of visual results, 
Fig. \ref{fig:exampleretrieve} shows several retrieved SAR patches using ResNet101-1024 that outputs 1024 dimensional feature vector and its backbone is ResNet101.
In most cases, the proposed method retrieves SAR images correctly from the database. 
In fact, only two retrieved images in the Exp.4 ykdelB-Hard case were incorrect; however, these images are quite similar with a query SAR image.

\section{Conclusions}\label{sec:conclusions}
In this paper, we have proposed a contrastive learning method for the SAR image retrieval task.
To this end, homography transformation is used for augmenting the SAR image patches and the output vector of a transformed image is compared against that of an original image.
This enables the proposed encoders to be trained in a self-supervised manner, thus no labelling process is necessary for establishing the dataset.
The proposed method is therefore especially applicable for tasks where the labelling process is time consuming.
Experiments were conducted and the results demonstrate that even in the cases where the encoders have not seen similar SAR images to a given query image, it nevertheless successfully retrieves similar SAR images from the database.
This work can be utilized for applications such as image searching based positioning or navigation, which is one our future extensions.

\bibliography{ref}
\bibliographystyle{unsrtnat}

\end{document}